%% file: main.tex
\documentclass[sn-vancouver,Numbered]{sn-jnl}% Math and Physical Sciences Numbered Reference Style 
\usepackage{graphicx}%
\usepackage{multirow}%
\usepackage{amsmath,amssymb,amsfonts}%
\usepackage{amsthm}%
\usepackage{mathrsfs}%
\usepackage[title]{appendix}%
\usepackage{xcolor}%
\usepackage{textcomp}%
\usepackage{manyfoot}%
\usepackage{algorithm}%
\usepackage{algorithmicx}%
\usepackage{algpseudocode}%
\usepackage{listings}%
\usepackage{ulem}%enables \sout{strikthrough text}
\usepackage{caption}%enables captions in equations 
\DeclareCaptionType{mycapequ}[][List of equations]
\usepackage{multirow}
\usepackage{caption}

\usepackage{tabularx}
\usepackage{longtable} % Tables that can span several pages
\usepackage{colortbl}
\usepackage{makecell}

\makeatletter
\usepackage{comment}
\let\wfs@comment@comment\comment
\let\comment\@undefined
\usepackage{changes}
\@namedef{Changes@AuthorColor}{red}
\definechangesauthor[color=red]{A}
\makeatother
\definechangesauthor[color=blue]{T}
\usepackage[shortlabels]{enumitem}

\theoremstyle{thmstyleone}%
\theoremstyle{thmstyletwo}%

\theoremstyle{thmstylethree}%

\RequirePackage{booktabs}

\usepackage{booktabs}\let\cline\cmidrule

\begin{document}

\title[Article Title]{
%TETYS: an LLM-based topic modeling pipeline for big-text datasets of scientific abstracts  
Capturing research literature attitude towards Sustainable Development Goals: an LLM-based topic modeling approach
}


\author[1]{\fnm{Francesco} \sur{Invernici}}\email{francesco.invernici@polimi.it}

\author[1]{\fnm{Francesca} \sur{Curati}}\email{francesca.curati@mail.polimi.it}

\author[1]{\fnm{Jelena} \sur{Jakimov}}\email{jelena.jakimov@mail.polimi.it}

\author[1]{\fnm{Amirhossein} \sur{Samavi}}\email{amirhossein.samavi@mail.polimi.it}

\author*[1]{\fnm{Anna} \sur{Bernasconi}}\email{anna.bernasconi@polimi.it}

\affil[1]{\orgdiv{Department of Electronics, Information and Bioengineering}, 
\orgname{Politecnico di Milano}, \orgaddress{\street{Via Ponzio 34/5}, \city{Milano}, \postcode{20133}, \country{Italy}}}

\abstract{

The world is facing a multitude of challenges that hinder the development of human civilization and the well-being of humanity on the planet.
The Sustainable Development Goals (SDGs) were formulated by the United Nations in 2015 to address these global challenges by 2030. 

Natural language processing techniques can help uncover discussions on SDGs within research literature. 
We propose a completely automated pipeline to 
1) fetch content from the Scopus database and prepare datasets dedicated to five groups of SDGs;
2) perform topic modeling, a statistical technique used to identify topics in large collections of textual data;
and 
3) enable topic exploration through keywords-based search and topic frequency time series extraction.

For topic modeling, we leverage the  stack of BERTopic scaled up to be applied on large corpora of textual documents (we find hundreds of topics on hundreds of thousands of documents), introducing
i) a novel LLM-based embeddings computation for representing scientific abstracts in the continuous space and
ii) a hyperparameter optimizer to efficiently find the best configuration for any new big datasets.
We additionally produce the visualization of results on interactive dashboards reporting topics' temporal evolution. Results are made inspectable and explorable, contributing to the interpretability of the topic modeling process.

%brief summary and potential implications
Our proposed LLM-based topic modeling pipeline for big-text datasets allows users to capture insights on the evolution of the attitude toward SDGs within scientific abstracts in the 2006-2023 time span.
All the results are reproducible by using our system;
the workflow can be generalized to be applied at any point in time to any big corpus of textual documents.
}


\maketitle

\section*{Introduction}
\label{sec:introduction}

Sustainable Development Goals (SDGs) are 17 United Nations' global objectives identified to address some of the biggest challenges of human civilization~\cite{sdgs}. 
These goals include issues such as gender equality and education, poverty and hunger, health, and climate change. Each goal is designed to address a specific issue or a set of strongly related issues; however, all goals should work together to create a better and more sustainable future for humanity.
We use keywords that describe SDGs as our point of access to a scientific literature landscape that is typically very vast and for which easy, flexible exploration is problematic. 
We access academic research outcomes through the Elsevier Scopus database, which stores a rich content of abstracts along with their metadata, via their RESTful API \cite{elsevier_academic_api}.

For the analysis, we follow an unsupervised statistical approach based on natural language processing, specifically focused on topic modeling~\cite{krause2006data}.
Unsupervised Topic Modeling is used to discover and analyze latent topics within a document, without leveraging pre-existing labels or supervision. 
This method works under the assumption that each document represents a single topic, or at least that one topic is preponderant, so as to exclude encompassing multiple topics at the same time.

In our work, we frame topic modeling as a clustering task~\cite{jayabharathy2011DocumentClusteringTopic} over the latent space generated by the LLM, differently from other approaches that build and train end-to-end models for topic modeling, both based on classical methods~\cite{moody2016MixingDirichletTopic} and on language models~\cite{meng2022TopicDiscoveryLatent}.
Egger and Yu~\cite{egger2022TopicModelingComparison} surveyed four topic modeling techniques, namely latent Dirichlet allocation, non-negative matrix factorization, Top2Vec, and BERTopic.
In line with their analysis and the suggestions of a more recent survey~\cite{abdelrazek2023topic}, we selected BERTopic~\cite{grootendorst2022bertopic} to implement our analyses, based on topic modeling from document clustering. 
Abdelrazek et al.~\cite{abdelrazek2023topic} confirms our hypotheses on the goodness of neural topic models for scalability -in terms of both model and data- and flexibility -i.e., the ability to adapt to different tasks like, in our case, dynamic topic modeling; these aspects are particularly important in our scenario.

BERTopic has already been successfully used for social sciences~\cite{falkenberg2022GrowingPolarizationClimate, ebeling2022AnalysisInfluencePolitical,scepanovic2023QuantifyingImpactPositive} since it is very flexible, can be scaled for big data corpora, and can be embedded in an end-to-end data pipeline.
We propose to use it in a different domain:
SDGs have triggered much interest as a key to understanding the general (both research and general public-driven) attitude toward high-stakes themes regarding transversally many continents and socioeconomic groups. 
Some work focused on extracting topics of discussion on social media comment threads~\cite{verma2024understanding,roldan2021understanding} or on online news~\cite{fitri2021topic}.
Saheb et al.~\cite{saheb2022artificial} targeted a small corpus of 182 research abstracts focused on a specific area
(artificial intelligence solutions for sustainable energy),
while 
Raman et al.~\cite{raman2024green} selected a small corpus of 448 research abstracts on green/sustainable AI.
Even if, to a small extent, the employed techniques and the domain of interest overlap with our interest, all mentioned works significantly differ from ours in the scale of their elaboration. 
Indeed, typically they are based on small datasets (a few hundred documents) and consequently build very small topic models (e.g., \cite{saheb2022artificial} identifies 8 topics, \cite{fitri2021topic} 10 topics, \cite{roldan2021understanding} 17 topics, and~\cite{raman2024green} 5 topics).
The work by Smith et al.~\cite{smith2023discovering} is more similar to ours in spirit; they analyze about 30k abstracts related to SDG 3 (Good Health and Well-being) and identify about 200 topics. 
Our innovation stands in making this kind of analysis completely reproducible on any big dataset and exposing it on a user-friendly interface.
Meanwhile, this allows us to complement previous efforts by providing a complete overview of all SDG-related keywords.

Here, we propose to adopt an LLM-based topic modeling pipeline named TETYS (standing for `Topics Evolution That You See'), which has the following characteristics:
\begin{itemize}
\item it can be run on big-text datasets in a completely automated mode;
\item it enhances BERTopic~\cite{grootendorst2022bertopic} default configuration with an LLM-based embedding computation;
\item it employs an innovative parameters' optimization mechanism that randomly searches the parameters' space to optimize a Density-Based Clustering Validation (DBCV) score -- thus making running the same pipeline on multiple big datasets feasible;
\item it allows us to build interpretable topic models for big corpora of complex (i.e., scientific/technical) text documents.
\item it builds a Web platform providing a complete overview of the topics, with interactive exploration of topics' representation over time.
\end{itemize}
In this manuscript, we deliver the results of applying TETYS on five groups of documents (called macro-areas)
that concern a collection of SDGs-related keywords (respectively on Basic Human Needs and Well-being; Environmental Sustainability; Economic Development and Employment; Equality and Social Inclusion; and Global Partnerships and Peace).
The pipeline was optimized to be run on each such group of documents. Our TETYS platform %, exposed at 
%\url{https://geco.deib.polimi.it/tetys/},
is a Web interface that makes results explorable to any stakeholder.

\section*{Materials and Methods}
\label{sec:materials}

We overview the preparation of the text corpora used for the analysis, then describe the TETYS pipeline, divided into its sub-pipeline for building and fitting the topic model and its sub-pipeline dedicated to topic exploration artifacts.

\subsection*{Datasets preparation}

We extracted all publications from Scopus, one of the largest repositories for academic abstracts and citations of peer-reviewed documents, including journal articles and conference proceedings. 
Scopus was established by the academic publisher Elsevier~\cite{elsevier} and is considered relatively more comprehensive than Web of Science~\cite{mongeon2016journal}. 
Scopus has enabled many text mining approaches, also using topic modeling~\cite{robledo2022topic} on very specific domains such as personal information privacy~\cite{choi2017analyzing} or public procurement~\cite{rejeb2023landscape}.

Next, we detail how we grouped the SDGs to define five overarching macro-areas that include a significant number of abstracts to be analyzed with our approach.
Then, we describe the strategy to retrieve abstracts and their metadata from Scopus API and, finally, we detail the data cleaning process.

\subsubsection*{Definition of SDG macro-areas}

We grouped the initial SDGs into macro-areas so as to make it easier to identify big topics, trends, and relationships, thereby providing a clearer big picture of sustainable development as a whole.
We chose not to exceed some hundred thousand documents, as this proved effective in previous works~\cite{invernici2024exploring} and it is recommended in BERTopic documentation~\cite{githubbertopic}.

We queried ChatGPT~\cite{chatgpt} with an appropriately crafted prompt asking to group the 17 SDGs into 5 macro-areas, each concisely described through 7 keywords, which are likely to be selected by the authors of the scientific papers (see Table~\ref{tab:groups}).

\begin{table}
\makebox[\textwidth][c]{
\begin{tabular}{p{0.4cm} p{5.8cm} p{6.1cm} p{1.5cm}}
\textbf{MA} & \textbf{Included SDGs} & \textbf{Keywords} & \textbf{\#abst.} \\
\midrule

\multirow{5}{*}{M1} & 1 No Poverty & 
\multirow{5}{=}{Poverty alleviation; Food security; Public health; Education access; Water quality; Sanitation infrastructure; Healthcare provision.} & 333,901 \\
& 2 Zero Hunger &&(original)\\ %$\rightarrow$ 333,901\\
& 3 Good Health and Well-being && \textbf{320,798} \\
& 4 Quality Education &&\textbf{(final)}\\
& 6 Clean Water and Sanitation &\\
\midrule

\multirow{6}{*}{M2} & 7 Affordable and Clean Energy & 
\multirow{6}{=}{Renewable energy; Urban sustainability; Sustainable consumption; Climate change mitigation; Marine biodiversity; Ecosystem conservation; Energy efficiency.} & 399,922 \\
& 11 Sustainable Cities and Communities &&(original)\\ % $\rightarrow$ 368,284\\
& 12 Responsible Consumption and Production && \textbf{339,949} \\
& 13 Climate Action &&\textbf{(final)}\\
& 14 Life Below Water \\
& 15 Life on Land \\
\midrule

\multirow{4}{*}{M3} 
& 8 Decent Work and Economic Growth &
\multirow{4}{=}{Economic growth; Innovation ecosystems; Infrastructure development; Entrepreneurship support; Industrialization strategies; Industrial Innovation; Labor market dynamics.} & 50,482\\
& 9 Industry, Innovation, and Infrastructure &&(original)\\ %$\rightarrow$ 45,116\\
&&&\textbf{41,218}\\
&&&\textbf{(final)}\\
\midrule

\multirow{4}{*}{M4} 
& 5 Gender Equality &
\multirow{3}{=}{Gender empowerment; Social equity; Inclusive policies; Women's rights; Minority rights; Income inequality; Social justice.} & 33,769 \\
& 10 Reduced Inequality && (original)\\ %$\rightarrow$ 31,380\\
&&& \textbf{25,017}\\
&&& \textbf{(final)}\\
\midrule

\multirow{4}{*}{M5} & 16 Peace, Justice, and Strong Institutions &
\multirow{4}{=}{Legal institutions; International cooperation; Peace efforts; Sustainable development cooperation; Global governance; Justice systems; Multilateral agreements.} & 56,275 \\
& 17 Partnerships for the Goals && (original)\\ %$\rightarrow$ 48,287\\
&&& \textbf{33,769} \\
&&& \textbf{(final)}\\
\bottomrule
\end{tabular}
}
\caption{Description of five macro-areas (MA) grouping the SDGs. 
M1 = Basic Human Needs and Well-being;
M2 = Environmental Sustainability; 
M3 = Economic Development and Employment;
M4 = Equality and Social Inclusion; 
M5 = Global Partnerships and Peace.
Numbers of abstracts are reported as 
i) number of original abstracts, and 
ii) number of abstracts after deduplication and data cleaning (in bold type).}
\label{tab:groups}
\end{table}
%https://tex.stackexchange.com/questions/503792/multirow-in-tabularx (mi ha salvato per tabularx+multirow

\subsubsection*{Data and metadata retrieval}

Due to its extensive coverage and being one of the most trusted databases in the academic field, we deemed Scopus to be an interesting source for our purpose.
Scopus evaluates the quality of its journals annually using four numerical measures: the h-Index, CiteScore, SCImago Journal Rank, and Source Normalized Impact per Paper. This review process ensures that the journals listed in Scopus meet the peer review quality standards required by several research grant agencies, as well as by degree-accreditation boards in many countries. 
Scopus covers 240 disciplines providing over 94 million records, including articles, conference papers, patents, and book chapters, in more than 105 countries. 

We accessed programmatically the corpus of literature data provided by Scopus, by employing its APIs~\cite{elsevier_academic_api} through two endpoints:
\begin{enumerate}
\item[(1)] Scopus Search API~\cite{scopus_search_api}. This search resource enables users to submit queries to the Scopus index and retrieve relevant metadata in user-specific text formats and the link to the corresponding Scopus abstracts.
\item[(2)] Abstract Retrieval API~\cite{scopus_retrieval_api}. This interface allows us to retrieve a Scopus abstract after searching the text of abstracts using the Search Scopus API.
\end{enumerate}
%Abstracts can be searched with Search Scopus API, and after finding relevant abstracts, those can be retrieved using the Scopus Abstract Retrieval API.
The endpoint (1) %provides a RESTful service at the address \texttt{https://api.elsevier.com/content/search/scopus}. %, for searching against the SCOPUS cluster, which contains SCOPUS abstracts. 
uses a \texttt{query} parameter that allows a boolean search with field restriction;
we employ the fields 
\texttt{pubstage} set to ``final'' to exclude preprints, 
\texttt{pubyear} starting from 2006 up to 2023 included, 
\texttt{language} to include English-language abstracts, and \texttt{key} for specifying keywords related to the abstracts (contained in author-specified keywords or automatically-indexed keywords).
By enclosing terms to be searched in double quotation marks, we employ a similarity-based
``search for a loose or approximate phrase'' exposed by Elsevier API~\cite{elsevierdev}. 
%https://dev.elsevier.com/sc_search_tips.html
%so as to find documents where the search terms are present in the specified fields, possibly also located near each other, rather than being spread apart within the text. 
While (1) fetches the identifiers of documents of interest, the actual abstracts with their metadata are retrieved by calling (2), one paper at a time. Scopus imposes a weekly limitation on the number of requests, resulting in a time-consuming process.
%For a total of 874,349 abstracts across all macro-areas, with an increased weekly quota of 100,000 abstracts, it took approximately 9 weeks to complete the data collection process in 9 batches.

\subsubsection*{Data cleaning}
\begin{figure}[h!]
    \centering
    \includegraphics[width=0.55\linewidth]{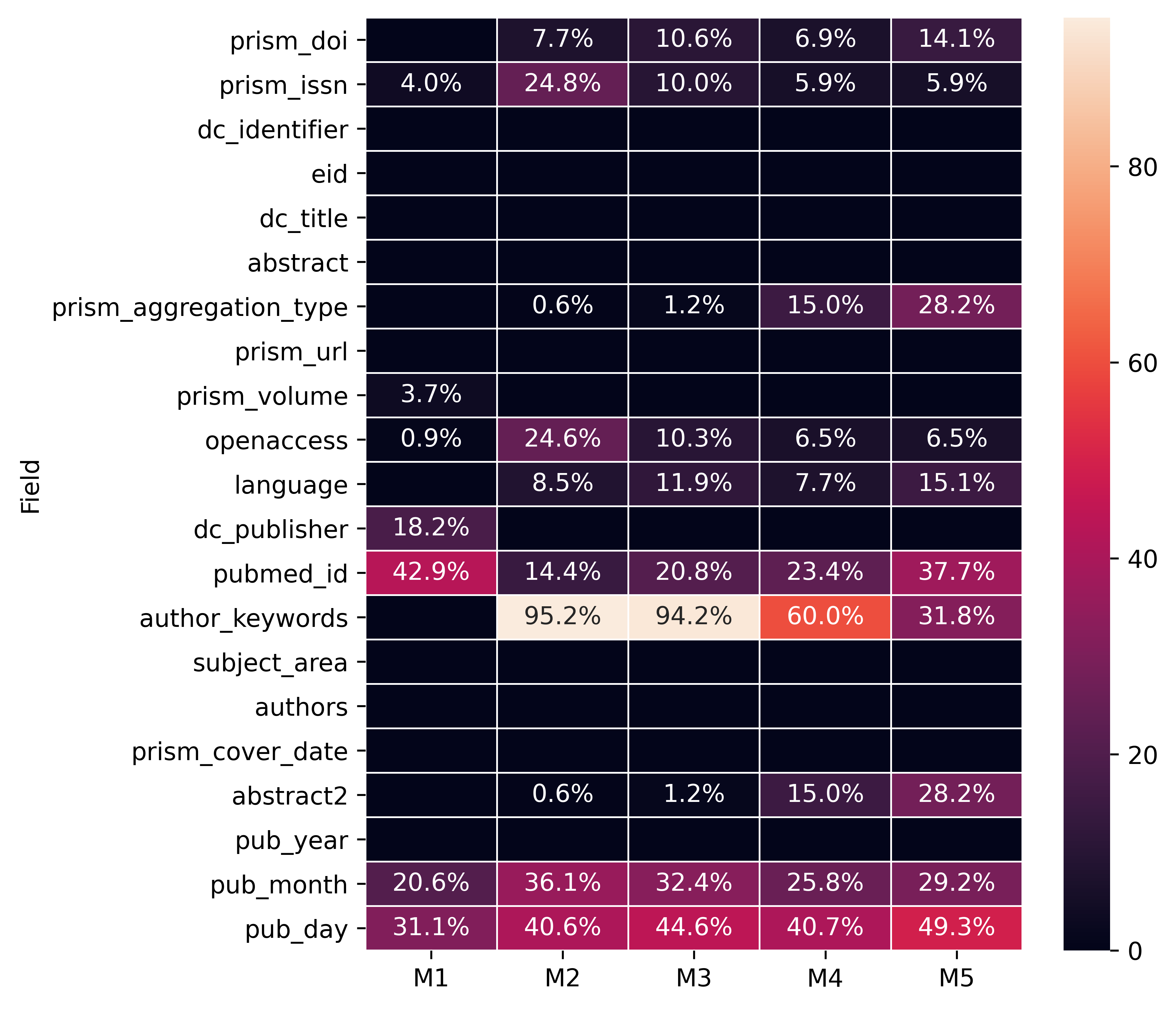}
    \caption{Heatmap representing the percentage of missing metadata API fields (rows) per macro-area (columns). Cells with no number indicate that the metadata field is present in all records. Lighter colors indicate the metadata field is heavily lacking.}    \label{fig:missing-values}
\end{figure}

For each macro-area, we obtained a dataset of ten-to-hundred thousands of documents (see numbers in the last column of Table~\ref{tab:groups}), each equipped with a set of 20 metadata fields. 
%\textit{doi}, \textit{issn}, \textit{coverDate}, \textit{dcIdentifier}, \textit{eId}, \textit{dcTitle}, \textit{abstract}, \textit{aggregationType}, \textit{url}, \textit{volume}, \textit{openaccess}, \textit{language}, \textit{dcPublisher}, \textit{pubmedId}, \textit{authorKeywords}, \textit{subjectArea}, \textit{authors}, \textit{pubYear}, \textit{pubMonth}, and \textit{pubDay}.
We removed from the metadata set the rows that did not have a corresponding abstract document, or that lacked a Digital Object Identifier (DOI), title, or publication date (see Figure~\ref{fig:missing-values} for the distribution of missing values per each metadata field).

%A DOI is a persistent ISO identifier that uniquely identifies different objects~\cite{doi}; we assume that two identical papers have the same \texttt{doi} parameter in both databases.
Then, we performed data deduplication for rows with the same digital object identifier and/or internal Scopus identifier.
Finally, we enforced the time window of interest for the publication date, keeping only abstracts published between 2006 and 2023 (included), and converted the dates into the Python DateTime format. At the end of the stage, we enforced the selection of abstracts written in English. 
Refer again to Table~\ref{tab:groups} (last column, second value) for counts of papers after the deduplication.

\begin{figure}[h!]
    \centering
    \includegraphics[width=\linewidth]{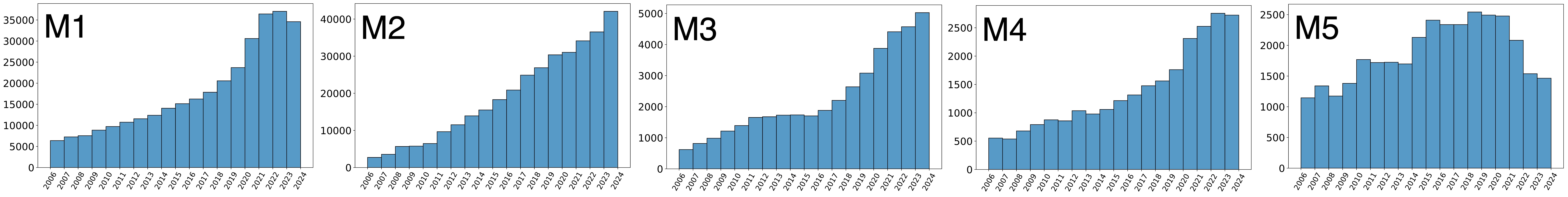}
    \caption{Data distribution over the years for all five macro-areas.}
    \label{fig:new_data_distribution}
\end{figure}

In Figure~\ref{fig:new_data_distribution}, we present the distribution of abstracts published each year, in the considered period, for each macro-area (M1 to M5).
The trend shows a general increase, which confirms aspects such as the increased global awareness of sustainability issues, the development of technology, and the growing number of researchers.
Interestingly, M1 (Basic Human Needs and Well-being) and M4 (Equality and Social Inclusion) show a spike during the period 2020-2023, likely due to the COVID-19 pandemic, while M5 (Global Partnerships and Peace) exhibits a less right-skewed distribution w.r.t. others.

\subsection*{TETYS Pipeline}

\begin{figure}[h!]
    \centering    \includegraphics[width=\linewidth]{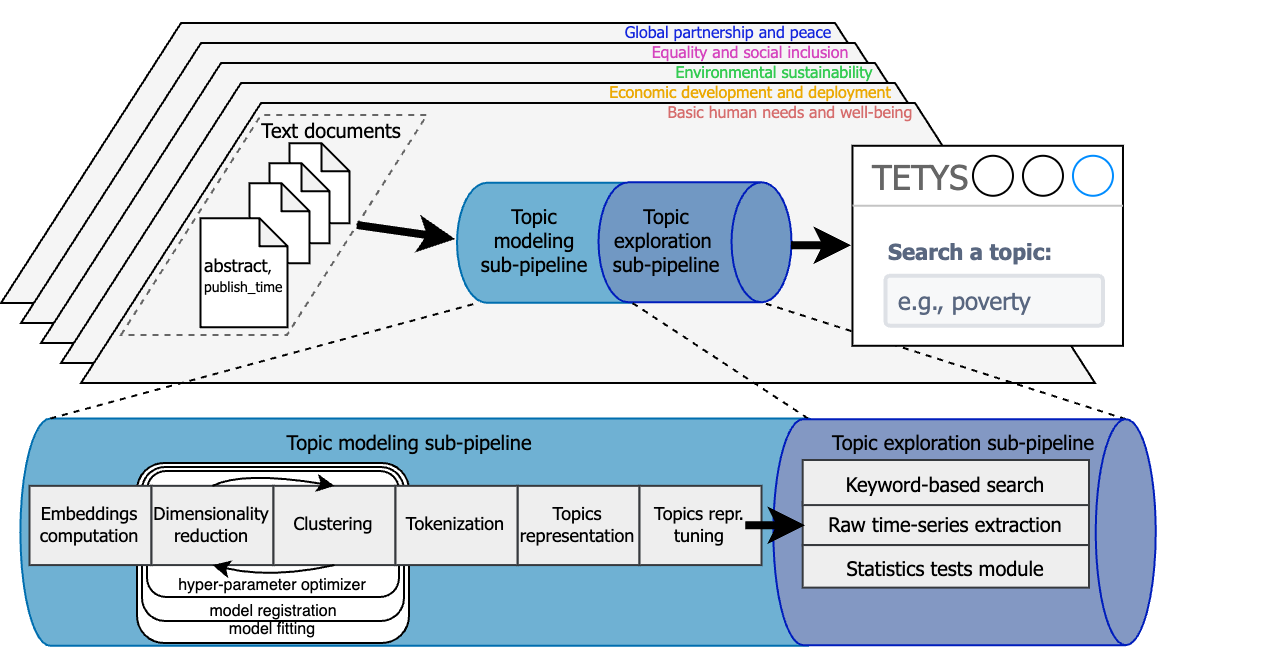}
    \caption{TETYS pipeline architecture.}
    \label{fig:pipeline}
\end{figure}

Our pipeline consists of two sub-pipelines (see Figure~\ref{fig:pipeline}), one dedicated to topic modeling and one to topic exploration.
The first sub-pipeline is dedicated to building a solid topic model and fitting it to the current dataset, arranging for an interpretable model representation.
The second sub-pipeline is concerned with extending the information within the topic model, allowing exploration via keyword-based search and adding simple distance metrics and time series on which statistical tests can be drawn. 

Every step in the two sub-pipelines is performed on five different datasets (each based on one of the previously defined macro-areas); each process produces, as a result, a topic model and auxiliary data structures that can be explored in a Web-based dashboard.
The pipeline instances are completely separated; when appropriate, others could be generated independently one from the other as the data architecture, the backend, and the frontend are general and can be configured based on need.

\subsubsection*{Topic modeling}

We base our work on BERTopic~\cite{grootendorst2022bertopic}, a topic modeling framework that leverages six steps to achieve unsupervised latent topic identification and textual representation learning. It requires
1) converting documents into embeddings,
2) reducing the dimensionality of the embeddings;
3) clustering the reduced embeddings;
4) tokenizing documents;
5) using a word-weighting scheme; and 
6) optionally tuning the obtained topic representation.

The default configuration employs, respectively, in the first five steps:
the sentence-transformer BERT (SBERT~\cite{reimers2019SentenceBERTSentenceEmbeddings});
the Uniform Manifold Approximation and Projection (UMAP) dimension reduction technique~\cite{mcinnes2020UMAPUniformManifold};
the Hierarchical Density-Based Spatial Clustering of Applications with Noise (HDBSCAN)~\cite{mcinnes2017HdbscanHierarchicalDensity}; 
the word tokenizer CountVectorizer~\cite{pedregosaScikitlearnMachineLearning};
and a class-based term frequency–inverse document frequency (c-TF-IDF) model~\cite{ceri2013InformationRetrievalModels}.

Since BERTopic's first conception, several enhancements have been introduced. Thanks to its modular structure and the possibility of completely customizing its pipeline, we searched for the best possible configuration given each macro-area domain and dataset at hand.
%adopted some of the improvements that have been proposed and tested our own customizations. 
With respect to a standard configuration of the BERTopic stack, TETYS introduces several contributions:
\begin{itemize}
\item we replaced the default SBERT with a Large Language Model (LLM) for the computation of embeddings;
\item we designed an innovative systematic \textit{optimizer} for the two hyperparameter-tuning steps of the pipeline (dimensionality reduction of embeddings and their clustering) -- this mechanism allows us to evaluate multiple configurations with different parameters, quickly converging to a (local) optimal one;
\item we implemented a model \textit{registration} functionality, to persist the output of the optimization phase and the consequent model fitting.
\end{itemize}
In the following, we discuss more in-depth these three novelties, followed by a brief description of the classical steps offered by BERTopic (including the tokenization and the representation of topics with its tuning).

\paragraph*{LLM-based embeddings computation}

In order to learn the latent topic structure of the dataset, we map each abstract in our datasets to a point in an embedding representation.

On June 20th, 2024, we inspected the
Massive Text Embedding Benchmark (MTEB) leaderboard~\cite{mteb} and selected the general-purpose model that maximized the average performance over a set of criteria listed by the leaderboard, while satisfying the memory constraints of our setup.
We employ a virtual machine equipped with an NVIDIA A100 (40GB) GPU~\cite{a100}, 32 virtual CPUs, 64 GB RAM, 60 GB SSD, and 500 GB HDD.

We selected the second release of the Salesforce embedding model (SFR-Embedding-2 R LLM \cite{SFR-embedding-2}).
The model was trained on abstracts concatenated with the corresponding paper title, producing 4096-dimensional embedding representations.
This choice replaced the default component SBERT proposed in~\cite{grootendorst2022bertopic} (with featured a much lower dimensional space).
The selected model is known to bring enhancements across all downstream tasks, with particularly notable improvements in clustering and classification tasks, making it a top-performance model on the HuggingFace MTEB benchmark leaderboard, at the time of our development. 

In the absence of documentation for the SFR-Embedding-2 R model, we referred to the SFR-Embedding-Mistral~\cite{sfr-embedding-mistral} model, its closest documented ancestor model. This is trained on a variety of data from different tasks. For clustering tasks, it utilizes data sourced from the preprint repositories arXiv, bioRxiv, and medRxiv, while applying filters to exclude development and testing sets.

SFR-Embedding-2 R, with $>$ 7 billion parameters, was used to run both training and fitting tasks; in our setup, its instance occupies nearly 27 GB when loaded into the GPU memory (out of  40 GB available) and it uses 26.49 GB (fp32) memory (out of 64 GB RAM available).

Loading the SFR-Embedding-2 R model and dataset into GPU memory was non-trivial. Due to its large size, it was impossible to simultaneously load the model and dataset and encode the abstracts into embedding vectors. We exploited the \texttt{transformers.pipelines} API~\cite{pipelines} and its built-in mechanisms for lazy loading and on-demand processing, which efficiently manage memory usage. The pipeline processes the data in manageable chunks, not requiring the whole data to be loaded in the GPU memory, only the necessary parts of the model and data are loaded when needed.

%The main three approaches used in the development of the SFR-Embedding-Mistral model are Transfer Learning from Multiple Tasks, \added{FORSE NON RILEVANTE?}
%The authors of the model have noticed that the embedding model performs much better in retrieval tasks when it is used alongside clustering tasks and it can be further improved through transfer learning from different tasks.
%Task-Homogeneous Batching, \added{FORSE PIU RILEVANTE?} 
%This technique creates batches made entirely of samples from one specific task. When applied to contrastive learning, as a result, the negative examples within each batch become more difficult to distinguish. This can force the model to learn more discriminating features to distinguish between similar items, which can lead to better generalization.
%Constructing Better Hard Negatives \added{FORSE PIU RILEVANTE?} 
%Hard negatives are data points that are particularly challenging for the model to distinguish from positive data points. This approach aims to create and select negative examples that are most similar to the positive ones, making them harder to differentiate. Improving the training of hard negatives sharpens the model’s ability to distinguish between closely related but misleading documents. 

\paragraph*{Hyperparameter optimizer}
In order to evaluate the goodness of the intermediate topic models that are generated (each one based on a specific configuration of the parameters set), we introduce an optimization mechanism.
In our previous work~\cite{invernici2024exploring}, we had proposed to optimize hyperparameters by performing a grid search, i.e., trying all the possible combinations to maximize the clusters' one-to-one relative density connection using the Density-Based Clustering Validation (DBCV)~\cite{moulavi2014DensityBasedClusteringValidation} index (spanning -1 for lowest quality to 1 for highest quality).
%DBCV uses the concept of density-contour trees, inspired by Hartigan’s model~\cite{hartigan1977distribution} to assess the quality of clustering~\cite{moulavi2014density}. 
%Hartigan’s model of Density Contour Trees defines \textit{density-based clusters} as regions of high density that are separated from other such regions by areas of low density. 
%The density within a cluster reflects its internal cohesion, with higher density indicating a more tightly connected cluster. In contrast to within-cluster density, the density between clusters determines their separation, with lower density suggesting that the clusters are well-separated and distinct from one another, possibly non-overlapping.
%According to Hartigan’s model, a strong density-based clustering solution should find clusters where the least dense region within each cluster remains denser than the most dense region in the areas separating the clusters~\cite{moulavi2014density}.
The DBCV score is a performance metric for clustering algorithms; however, we leveraged this metric for all our hyperparameters as DBCV not only assesses the quality of the clusters but also provides valuable insights into the cohesiveness and separation of topics.

Clearly, with grid search, we can always achieve the optimal configuration, even if at the cost of spending a significantly longer time. 
Here, we experiment with random search, which involves sampling a fixed number of hyperparameter combinations (much smaller than the total number of possible configurations).
With this option, we obtain satisfactory results, allowing us to scale our approach up to any number of TETYS execution pipelines; specifically, we propose the following steps:
\begin{enumerate}[label=(\arabic*)]
\item We generate the parameters' space including four parameters for dimensionality reduction (UMAP) and four parameters for clustering the embeddings (see Table~\ref{tab:parametersranges} for the parameters ranges including the tested $\langle$ start, end, step $\rangle$ scheme).
\item We define a finite number of random search steps (empirically, we appreciated that --once around the 100th step-- the local maximum solution found by the random search typically resembles the global maximum one found with the grid search approach).
\item Until the number of steps identified in (2) is not reached, we experiment with one configuration at a time as follows:

\begin{enumerate}[label=(\roman*)]
\item Draw one configuration in the parameters' space (see Table~\ref{tab:parametersranges}). %with no repetition
\item Run UMAP and HDBSCAN with the selected configuration on a validation subset of the current dataset (a randomly sampled 20\% of the dataset).
\item Calculate the corresponding DBCV score.
\item If the DBCV score \textbf{is not} the current best (local) maximum one, discard the configuration and proceed to the next one. If \textbf{it is} the current best one, proceed with \textit{Model registration} and \textit{Model fitting}.
\end{enumerate}
\item The model with the highest DBCV (once the random search steps are concluded) is considered the best one and employed for the following BERTopic steps.
\end{enumerate}

\begin{table}[h!]
    \centering
    \begin{tabular}{lllccccc}
        \textbf{Step} & \textbf{Parameter name} & \textbf{Parameter range}&\textbf{M1}&\textbf{M2}&\textbf{M3}&\textbf{M4}&\textbf{M5}\\        
        \midrule
        \multirow{4}{*}{UMAP} & n\_neighbors & (1, 100, 5) & 20&{20}&100&50&100\\
             & min\_dist & (0, 1, 0.05) & 0&0&0&0&0\\
             & n\_components & (5, 50, 5) & 5&10&10&28&35\\
             & metric & (`euclidean') &\multicolumn{5}{c}{--- `euclidean' ---}\\ %-> gli embedding sono normalizzati, quindi euclidean è uguale a cosine, ma veloce da calcolare
             
        \midrule
        \multirow{3}{*}{HDBSCAN} & min\_samples & (10, 100, 10) & 75&75&10&10&15\\
                & min\_cluster\_size & (25, 100, 5) & 25&25&25&{25}&25\\
                & cluster\_selection\_method & (`eom', `leaf') &\multicolumn{5}{c}{--- `eom' ---}\\
        \bottomrule
    \end{tabular}
    \caption{For each step and parameter, we report the value ranges ($\langle$ start, end, step $\rangle$) tested by the \textit{optimizer} of the hyperparameters of the dimensionality reduction and the clustering steps. The last five columns report, for each of the five macro-areas, which parameters configuration led to the best DBCV performance, thus used for the model fitting.}
    \label{tab:parametersranges}
\end{table}

In the best run for each macro-area, we obtained DBCV scores of, respectively, 0.52, 0.76, 0.39, 0.46, and 0.38 using the parameters' values reported in the last five columns of Table~\ref{tab:parametersranges}.

\paragraph*{Model fitting and registration}

Once the optimizer has selected the final parameters set, we run the \textit{Model registration} and \textit{Model fitting} components.

During \textit{Model registration} we save the model in two formats:
(i) \texttt{pickle}, a binary object for quality checks during this optimization process;
(ii) \texttt{safetensors}, a PyTorch model~\cite{pytorch} ready to be used for future inference on new data that the model has not seen.
This component is designed to address the challenges posed by the stochastic nature of the HDBSCAN algorithm. It ensures that the best model found during hyperparameter optimization is saved immediately and preserved for future use.
The main advantage comes from the fact that reinitializing the BERTopic model, even with the same hyperparameters, can yield different results, due to the randomness involved in HDBSCAN initialization. This variability can lead to a model that underperforms compared to the one identified during the hyperparameter optimization phase. By incorporating the \textit{registration} component, we not only ensure that the integrity of the best-performing model is preserved, but also that any subsequent analysis or application of the model is based on a consistent and reproducible version.
A disadvantage of this approach is that we need to store multiple models; %during the hyperparameter optimization phase. 
since we do not know in advance which model will perform the best among all the models found, we need to keep track of several versions, consistently increasing memory usage. 
Additionally, the time required to fit the model can be significant for certain parameter configurations. To address this issue and avoid saving models that will not be useful, we added the possibility of saving the model only if
1) it is the best model found thus far, and 
2) its DBCV score is greater than a 0.30 threshold limit, which we identified empirically through manual inspection of preliminary results.

Then, \textit{Model fitting} involves exploiting the hyperparameters corresponding to the current DBCV score.
With these parameters, we run UMAP and HDBSCAN on the whole dataset (100\%).
Note that, while in UMAP the parameters correspond to hyperparameters observed during validation, 
for clustering we need to fit the model --with its selected hyperparameters-- to the data and compute the actual parameters (e.g., number of clusters, center of clusters, etc.).
As an outcome of running this component, we build the final models, on which subsequent steps of BERTopic are applied.

\paragraph*{Topic representation and tuning}

The three remaining steps in the BERTopic pipeline contribute to achieving interpretable, synthetic representations of topics.
The first step involves an abstract vectorization (performed with the default  \texttt{scikit-learn}~\cite{pedregosaScikitlearnMachineLearning} CountVectorizer).

Second, 
%Specifically, we set \texttt{stop\_words} to ``English'' and \texttt{token\_pattern} as a regular expression to keep together hyphenated words, such as COVID-19 and SARS-CoV-2, which are common in biomedical writings. 
we fit the c-TF-IDF model with the \texttt{reduce\_frequent\_words} parameter set, which considers the square root of the normalized frequency of the terms (i.e., words). With this model, we obtain the most relevant terms (i.e., topics) per class, with their frequency. 
This corresponds to a textual, human-understandable representation for each cluster.
The most important topics can be retrieved using the TF-IDF representations.

Third, to improve our topic representation, we target the reduction of similar keyword repetition, such as those with the same root word or variations (e.g., singular and plural forms of the same word). 
More distinct and meaningful keywords, without redundancy, ensure that each keyword adds value to the overall representation and understanding of the topic. 
The most suitable algorithm for our purposes is Maximal Marginal Relevance (MMR). MMR selects keywords for topic representation, based on their relevance score and their dissimilarity to previously selected items. The goal is to maximize the relevance score while minimizing redundancy.
MMR allows us to reduce redundancy and get a clearer, more accurate picture of the keywords, where topics are more distinct and meaningful while making them easier to understand and interpret.

\subsubsection*{Topic exploration}
While the first sub-pipeline essentially allows us to systematize the customization of a BERTopic-like process, the second sub-pipeline creates a set of support data structures and representations useful to make topic exploration possible on dedicated visual dashboards.

First, we adopt the \texttt{word\_cloud}~\cite{mueller2023Wordcloud} package to generate word clouds with the most frequent terms of each topic, thereby providing a visual representation to inspect the topic content.

Second, we enable a keyword-based search, by exploiting the \texttt{find\_topics} function implementation in BERTopic~\cite{grootendorst2022bertopic},
%https://maartengr.github.io/BERTopic/getting_started/search/search.html
which essentially allows inputting a simple search term (possibly including spaces) to retrieve a list of similar topics equipped with their score of similarity w.r.t. the input term. 

Third, we compute per-topic time-series, representing the counts of papers published during the observed period 2006-2023. 
Our approach builds time-series using a parametric number of months in each \textit{time bin}.
For each abstract, we consider the date when it was published and the topic it belongs to; then, given a time granularity (1-month, 3-month, 6-month, or year), we compute bins corresponding to the requested timeframe.
As an output, we obtain tuples of the form
$\langle topic\_id, (bin\_id,start\_date), \#abstracts\_in\_bin \rangle$.
This method resembles the Dynamic Topic Modeling techniques proposed within BERTopic~\cite{grootendorst2022bertopic}.
Essentially, we add the run-time computation of features that are useful for analyzing time-series: 
i) binning;
ii) absolute/relative frequency (we normalized the count w.r.t. the number of total abstracts published in that bin); and
iii) ranking.
In this way, we can interpret the values as pointwise measures of the intensities of the topic, as other previous works on dynamic topic modeling~\cite{krause2006data}.
Taking advantage of these time-series, we generate line plots for the counts of abstracts per bin.

Finally, we implement two statistical tests. 
To check if the trend difference of \textit{two} periods of the same topic is significant, we use the non-parametric Kruskal-Wallis test~\cite{kruskal1952UseRanksOneCriterion}, typically employed for comparing sample medians (checking if two groups are sampled from the same population). 
The test produces a p-value, enabling the acceptance or rejection of the simple null hypothesis ``there is no significant difference in the topic representation in periods T1 versus T2'' (we adopt the library \texttt{SciPy.stats.kruskal}~\cite{2020SciPy-NMeth}).
We use the 5\% p-value as the threshold for significance; lower p-values allow the rejection of the null hypothesis~\cite{kruskal1952UseRanksOneCriterion}.
To check if the trend difference of \textit{multiple} periods of the same topic is significant, we apply Kruskal-Wallis to all intervals and verify if at least one interval is significantly different from the others. To understand which interval deviates from others we use the Dunn test~\cite{dunn1964multiple} with multiple testing corrections.
%http://geco.deib.polimi.it/tetys_api/docs#/default/request_test_single_topic_multiple_intervals_analysis_single_topic_multiple_intervals_post

\section*{Results}

In the following, we describe the five obtained topic models, evaluate them with those obtained using a baseline pipeline, and finally propose the topic exploration dashboard.

\subsection*{Extracted topics overview}

In the five macro-areas we found, respectively, 550 topics (Basic Human Needs and Well-being), 856 topics (Environmental Sustainability), 181 topics (Economic Development and Employment), 136 topics (Equality and Social Inclusion), and 167 topics (Global Partnerships and Peace).
The number of identified topics is roughly proportional to the number of abstracts for each macro-area (see Table~\ref{tab:groups}).
M1 and M2 are the biggest macro-areas, as they also include more Sustainable Development Goals compared to the M3-M5.

\begin{figure}[h!]
    \centering
    \includegraphics[width=\linewidth]{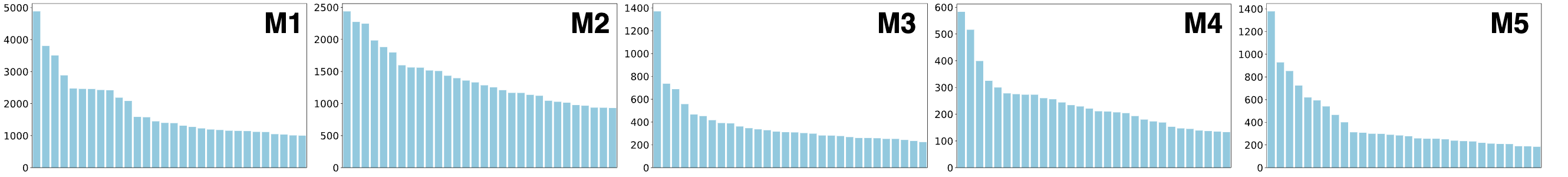}
    \caption{Distribution of the 30 largest topics based on the number of abstracts associated with each of them for each macro-area and configuration.}
    \label{fig:topic_distributions}
\end{figure}

For a quick overview, in Figure~\ref{fig:topic_distributions} we present diagrams illustrating the distribution of topics, only including the top 30 topics based on their abstracts' counts. The y-axis maximum values are 5,000 for M1, 2,500 for M2, 1,400 for M3, 600 for M4, and 1,400 for M5.

\begin{figure}[h!]
    \centering
    \includegraphics[width=\linewidth]{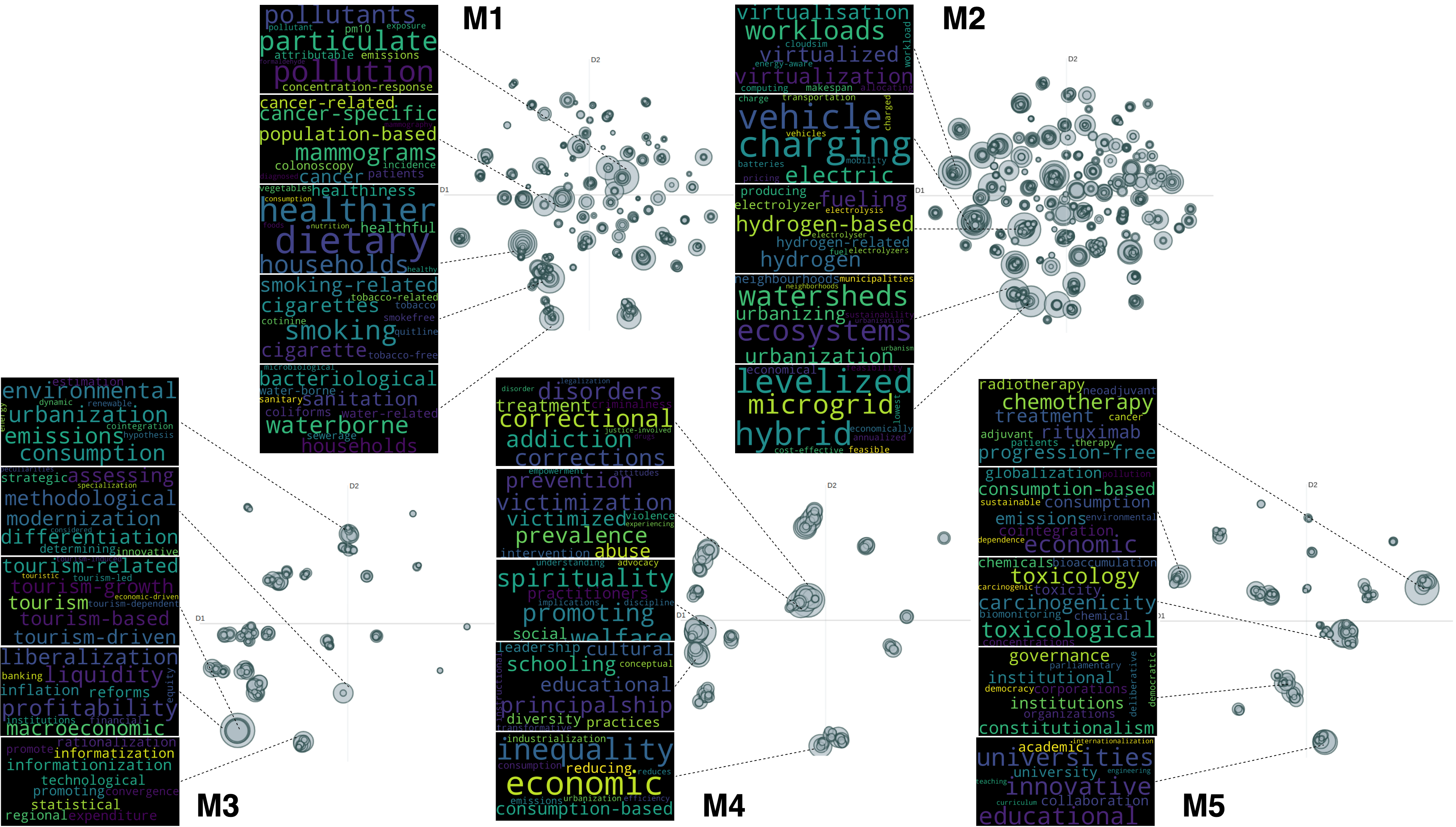}
    \caption{The biggest and most interesting topics from the five macro-areas}
    \label{fig:5macroareas}
\end{figure}

Figure~\ref{fig:5macroareas} shows, for each macro-area, its \textbf{intertopic distance map}.
This map places the topics in two dimensions, where the Euclidian distance between any two of them represents their similarity: the closer they are, the more semantically similar they are.
Topics are represented as circles and their size depends on the number of abstracts they gather. Due to the projection from a higher-dimensional space to two dimensions, we observe several overlaps in the map.
In the figure, the five largest topics for each area are connected to their corresponding word-clouds.

In M1 (Basic Human Needs and Well-being),
the `pollutants' and `bacteriological sanitation' topics are likely related to the \textit{Clean Water and Sanitation} goal (SDG 6).
Topics on `cancer' and `smoking' are closely connected with the \textit{Good Health and Well-being} goal (SDG 3).  
The topic related to `health and diets' is probably derived from publications related to \textit{Zero Hunger} (SGD 2); 

In M2 (Environment Sustainability),
the terms `workloads, virtualisation and energy-aware' seem related to the optimization of computing resources, and probably are in connection with energy consumption in data centers.
The `electric vehicle and charging' topic can also be related to the same goals. Hydrogen is considered a clean energy carrier~\cite{lubitz2007hydrogen} and is often connected with clean and renewable energy, thus, topics related to it can be connected to both \textit{Affordable and Clean Energy} and \textit{Responsible Consumption and Production goals} (SDGs 7 and 12). 
The topic with the `watersheds, urbanising and ecosystems' terms seems closely related to the \textit{Sustainable Cities and Communities} goal (SDG 11).  
The terms `levelized, microgrids, and hybrid' are often associated with sustainable energy problems and solutions, which are closely related to the \textit{Affordable and Clean Energy} goal (SDG 7). 

In M3 (Economic Development and Employment),
the topics on `liquidity, macroeconomic, and profitability' and on `tourism-related' are connected to the \textit{Decent Work and Economic Growth} goal (SDG 8).
Instead, the topics on `methodological modernization' and `environmental urbanization' appear related to the \textit{Industry, Innovation, and Infrastructure} goal (SDG 9).

In M4 (Equality and Social Inclusion),
the topic of `victimization and abuse' are related to the \textit{Gender Equality} goal (SDG 5), while other topics can be connected more generally to the \textit{Reduced Inequality} goal (SDG 10). 

In M5 (Global Partnerships and Peace),
topics look very versatile, possibly because the concept of ``partnership'' can include many different ideas and realizations.

\subsection*{Evaluation of topic modeling results}

We proposed a customized implementation of the BERTopic pipeline, where a local optimal configuration can be found by exploiting our hyperparameters optimization and model registration mechanisms.
This procedure is necessary due to the high quantity of data and the need to use many different models (e.g., five in our case) to be trained and fitted at the same time.
A quantitative evaluation of a model can be obtained at each single iteration (with a new candidate hyperparameter configuration) by leveraging the DBCV score. 
Then, the final selected configuration is assessed through a manual evaluation, as described next.

For evaluating the TETYS pipeline we compared two different configurations used in our specific use case:
\begin{itemize}
    \item \textbf{Baseline}: {Allenai-SPECTER} Embedding Model (this is a non-LLM model developed by AllenAI~\cite{cohan2020SPECTERDocumentlevelRepresentation}), with hyperparameters (exact) grid search method.
    \item \textbf{TETYS}: {SFR-Embedding 2 R} Embedding Model (\cite{SFR-embedding-2}), with hyperparameters random search method.
\end{itemize}
Note that the Baseline configuration leverages Scientific Paper Embeddings using Citation-informed TransformERs (SPECTER), a pre-defined model developed to learn general-purpose vector representations of scientific documents. It builds on the architecture of Transformer-based language models, in particular SciBERT~\cite{beltagy2019scibert}, an adaptation of the BERT model architecture~\cite{devlin2019bert} to the scientific domain. 
The model is trained on abstracts concatenated with the corresponding paper title; it produces 768-dimensional embedding representations.
This configuration is much smaller and faster to fine-tune, thus, we use a grid search strategy for hyperparameter tuning, to iterate over all combinations of parameters. The embedding model is specialized for scientific documents, which perfectly corresponds to our task. 

On the other hand, the TETYS configuration is the novel one proposed in this work, as described in the `Materials and Methods' section.
This configuration is larger and very time-consuming for the fitting process. Since we introduced model registration in the original pipeline --storing the best-performing model identified at any point-- it became impractical to try all possible combinations for models, as fitting some models for certain macro-areas can take a long time (i.e., approximately exceeding an hour). 
For this reason, a random search strategy was used to avoid excessive computation time. Due to this, we may not find the best possible model (only, one that achieves a local maximum of the DBCV score). 
Note that the embedding model is more general and optimized for a broader range of tasks (differently from SPECTER).
Supplementary Table 1, in the Appendix, presents the values of the hyperparameters for the best models obtained using the two configurations in the five macro-areas scenarios.

\subsubsection*{Quantitative assessment}

%Variano:
%dim.red. = PCA vs UMAP (sempre ottenuto valori dbcv piu bassi)
%model = specter vs mistral(LLM di salesforce)
%par.opt = grid search(sempre ottimo) vs random search(+efficiente)

The Density-Based Clustering Validation (DBCV) quantitatively evaluates the quality and diversity of topics. 
It provides an overall score that allows us to assess 
embeddings computation, hyperparameter search (for dimensionality reduction and clustering) providing one optimal choice for a given dataset (macro-area), embeddings model, and parameters' configuration.

\begin{figure}
  \begin{minipage}[b]{.45\linewidth}
    \centering
    \resizebox{\textwidth}{!}{
    \begin{tabular}{ccccc}
 &\multicolumn{2}{c}{Baseline}&\multicolumn{2}{c}{TETYS}\\
 \cmidrule(r){2-3}\cmidrule(l){4-5}
 &  \textbf{\#topics} & \textbf{DBCV} & \textbf{\#topics}  &\textbf{DBCV} \\ \midrule
\textbf{M1}   & 301   &0.44  &  550  &\textbf{0.52}    \\ 
\textbf{M2}   & 424   &0.72  &  856  &\textbf{0.76}   \\ 
\textbf{M3}   &  98   &0.36  &  181  &\textbf{0.39}    \\ 
\textbf{M4}   &  42   &0.44  &  136  &\textbf{0.46}    \\ 
\textbf{M5}   & 291   &0.37  &  167  &\textbf{0.38}   \\ 
\bottomrule
\end{tabular}}
    {\captionof{table}{Number of topics identified by the models for each macro-area/configuration and maximum DBCV score achieved.}\label{tab:config_compar}}
  \end{minipage}\hfill
  \begin{minipage}[b]{.45\linewidth}
    \centering
    \includegraphics[width=\linewidth]{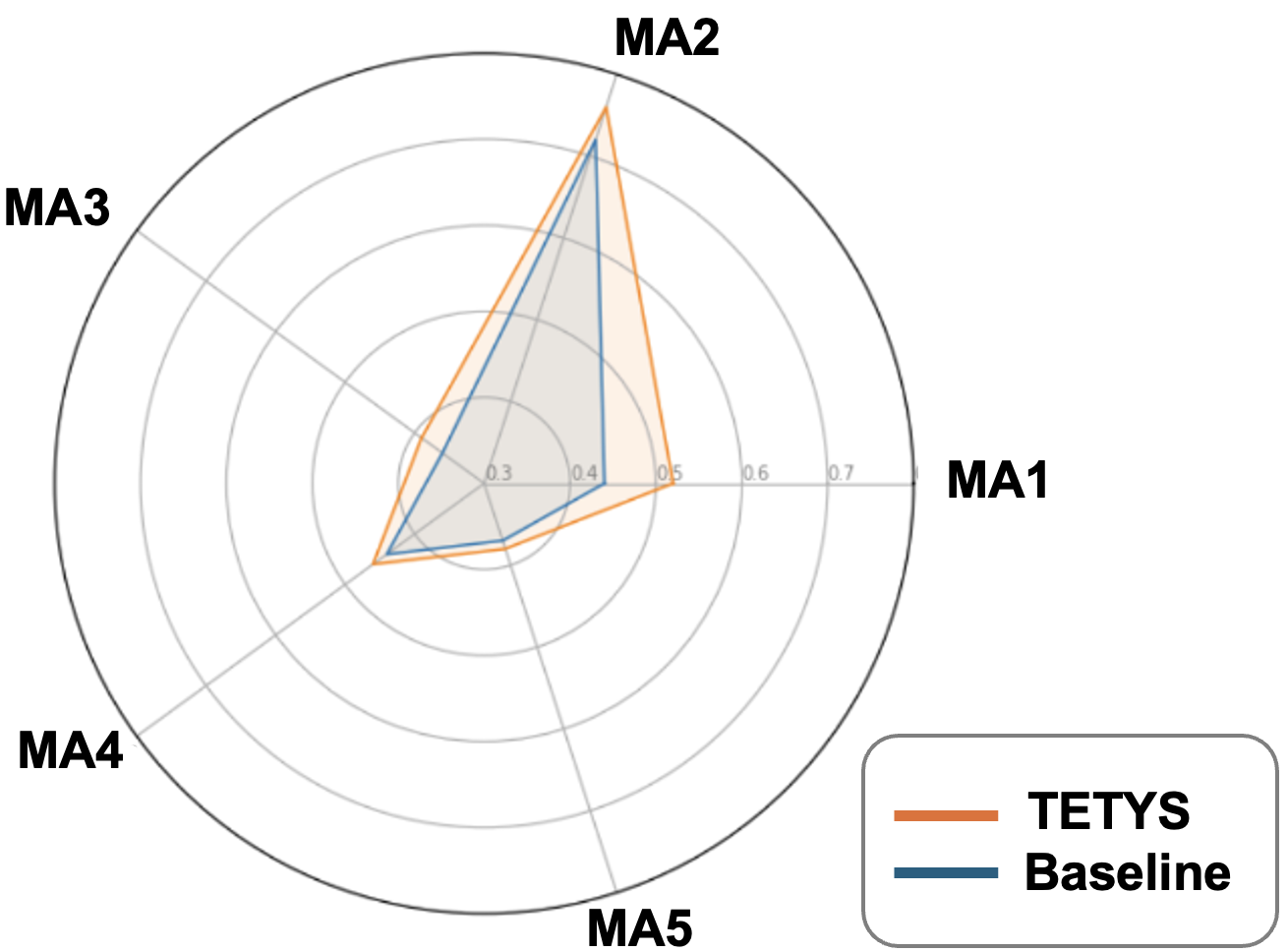}
    {\captionof{figure}{DBCV scores for both model configurations for five macro-areas}\label{fig:radar_plot}}
  \end{minipage}
\end{figure}

Table~\ref{tab:config_compar} provides an overview of the number of topics and corresponding DBCV scores for the five macro/areas.
DBCV scores are compared in the radar plot in Figure~\ref{fig:radar_plot}, showing an overall consistent improvement in the LLM-based configuration.

We note that for M1 (Basic Human Needs and Well-being) and M2 (Environment Sustainability), the LLM-based configuration model produced a significantly greater number of topics compared to the non-LLM-based configuration model. 
Surprisingly, for M5 --probably the most heterogeneous dataset (as observed in the analysis of the five largest topics of Figure~\ref{fig:5macroareas})-- the number of topics found with the Baseline configuration is consistently greater than the one with the TETYS configuration. This is possibly due to the particular combination of \texttt{n\_components} and the \texttt{n\_neighbors} parameter values in TETYS: we are using a higher dimensional space and forcing the model to look for a much larger neighborhood, resulting in fewer bigger clusters (w.r.t. the Baseline configuration).

\subsubsection*{Manual analysis}

%Then, we classify with both configs a set of 50 papers and manually check/assess (this is like looking at the wordcloud) the classification of the two configs.
%Report Precision, Recall, F1 for 
%G1, C1
%G1, C2
%G5, C1
%G5, C2

Topic modeling is an unsupervised technique,
which attempts to identify topics within a collection of documents without leveraging any other information, labels, or predefined topics. 
Since it is unsupervised, evaluating the quality of topic models becomes a challenging task that requires domain knowledge and expertise in the fields covered by the scientific papers under consideration. 

%To achieve the fairest possible evaluation, it would be necessary to read all the documents (i.e., papers' abstracts) involved in the training to gain a global understanding of how the model defined different topics. 
%Even in that case, the evaluation would be biased and dependent on the human evaluator. 

%In this thesis, our rationale for evaluating the performed work was to compare the topics identified by models with different configurations. 
%As we have already explained, the most important difference between the two configurations is the  LLM-based embedding models. 
The goal of our evaluation is to determine whether the LLM-based topic model is better at assigning topics (as we postulated), given the enhanced potential of the employed embedding model.
Our manual evaluation was carried out for two macro-areas, i.e., M1 (Basic Human Needs and Well-being) and M2 (Environment Sustainability), which are the largest ones and encompass the greatest number of Sustainable Development Goals.
 
We performed the inference on a test set of 100 abstracts for each macro-area; these abstracts were new, i.e., not seen by the models in the training phase (thus, here, we speak about `inference' rather than `fitting'). 
%They were obtained the same way as the training set, using the same keywords and database. 
In line with BERTopic, we assigned the special topic -1 to documents that do not belong to any valid topic, 
while topics labeled with numbers $[0, \#num\_topics - 1]$ are valid topics and are sorted from the largest to the smallest one.

%In Table \ref{tab:valid_topics_numbers} we present the number of abstracts with assigned valid topics for all four models used in the manual evaluation process.

%\begin{table}[h!]
%\centering
%\begin{tabular}{ccc}
%\textbf{Model} & \textbf{Total abstracts} & \textbf{Abstracts with valid topics ($\neq$ -1)} \\ 
%\midrule
%M1, Baseline & 100 & 56    \\ 
%M1, TETYS & 100 & 43      \\ 
%M2, Baseline & 100 & 53    \\ 
%M2, TETYS & 100 & 59      \\ 
%\bottomrule
%\end{tabular}
%\caption{Number of test abstracts with valid topics for the evaluated model configurations}
%\label{tab:valid_topics_numbers}
%\end{table}
Overall, we detected 56/100 abstracts with valid topics (i.e., $\neq$ -1) in M1 with the Baseline configuration, 43/100 in M1 with the TETYS configuration, 53/100 in M2 with Baseline, and 59/100 in M2 with TETYS.

After classifying the abstracts with both configurations, we asked two researchers who are experts respectively in the domains of M1 and M2 to manually assess each abstract.
%In Figure \ref{fig:manual_evaluation_input} we show one example of the file used as input for manual evaluation. 
%Columns "CORRECT" and "ALTERNATIVE" are filled out by the manual evaluator. 
%The "CORRECT" column shall indicate whether the predicted topic was correct, while the "ALTERNATIVE" column suggests, if appropriate, another appropriate topic.
They were equipped with a spreadsheet whose rows represent single articles; for each article, we provided the abstract, doi, and additional metadata (such as the author-defined keywords and the Scopus subject area).
For both the \textit{Baseline configuration} and the \textit{TETYS configuration} we provided the topic ID, topic probability, topic name, number of abstracts assigned to the topic, and the list of the ten most represented terms in the topic (along with their frequency).
Given this information, the evaluators were asked to 
indicate the identifiers of the most suitable topic among:
1) the ones available in the Baseline configuration; and
2) the ones available in the TETYS configuration.
By comparing the evaluators' choice with the ones derived from the automatic configurations, we computed the Precision, Recall, and F1-scores (see Table~\ref{tab:f1}).
TETYS achieves better results in all the indicators.
 
\begin{table}[h!]
\centering
%\resizebox{\textwidth}{!}{
\begin{tabular}{lccccccc}
&&\multicolumn{3}{c}{Baseline}&\multicolumn{3}{c}{TETYS}\\
\cmidrule(r){3-5}\cmidrule(l){6-8}
& \textbf{Avg type} & \textbf{Precision} & \textbf{Recall} & \textbf{F1} & \textbf{Precision} &  \textbf{Recall} & \textbf{F1} \\ \midrule
\multirow{3}{*}{M1} & Micro & 0.640 & 0.640 & 0.640 & 0.820 & 0.820 & 0.820 \\ 
& Macro & 0.569 & 0.540 & 0.547 & 0.668 & 0.654 & 0.658 \\ 
& Weighted & 0.612 & 0.640 & 0.601 & 0.737 & 0.820 & 0.768 \\
\midrule
\multirow{3}{*}{M2} & Micro & 0.690 & 0.690 & 0.690 & 0.870 & 0.870 & 0.870 \\
& Macro & 0.514 & 0.541 & 0.516 & 0.765 & 0.769 & 0.767 \\
& Weighted & 0.619 & 0.690 & 0.642 & 0.737 & 0.802 & 0.832 \\ 
 \bottomrule
\end{tabular}
%}
\caption{Precision, recall, and F1 scores for both configurations run on M1 and M2}
\label{tab:f1}
\end{table}
Moreover, we asked our evaluators to declare a preference between the assignment obtained using the Baseline configuration versus the one obtained using the TETYS configuration. Here, we allowed three possible choices:
\begin{itemize}
     \item the evaluator concludes that the assignment obtained by the \textbf{Baseline} configuration is superior;
     \item the evaluator concludes that the assignment obtained by the \textbf{TETYS} configuration is superior;
     \item none of the assignments is clearly superior w.r.t. the other one (\textbf{undefined}).
 \end{itemize}

\begin{table}[h!]
\centering
\begin{tabular}{ccc}
\textbf{Evaluator's choice} & \textbf{M1 Percentage} & \textbf{M2 Percentage} \\ \midrule
Baseline & 27\% & 21\%\\ 
TETYS & 32\% & 48\%\\ 
undefined & 41\% & 30\%\\ 
\midrule
\multirow{2}{*}{McNemar's test result} & p-value 0.6 & \textbf{p-value 0.001}\\
&statistic 27.0 & statistic 21.0\\
\bottomrule
\end{tabular}
\caption{Ballot comparison, with statistical evidence that TETYS configuration is strongly preferable to the Baseline in the case of M2.}
\label{tab:barlot_comparison}
\end{table}
Table \ref{tab:barlot_comparison} reports the number of each selected option in percentage.
We statistically tested the preference of one configuration over the other; along the guidelines indicated in Schuff et al.~\cite{schuff2023human}, we performed the non-parametric McNemar statistical test~\cite{mcnemar1947note} (used for paired nominal data), ignoring the `undefined' cases. 
For M1 (Basic Human Needs and Well-being) we observed  \textit{no statistical preference} between the two configurations; instead, for M2 (Environment Sustainability) we observed a \textit{strong statistical preference} for the TETYS, according to the low (0.001) obtained p-value. 

From this small experiment, we conclude that the LLM-based configuration is slightly better or at least as good as the non-LLM-based configuration. 
We expect that such restrained improvement is due to the use of the random search strategy for the LLM-based model, which means that we likely settled for a model that is not the best possible one.

By manual inspection of topics, we also observed that the TETYS configuration allowed us to achieve better quality, interpretability, and diversity~\cite{abdelrazek2023topic}. 
TETYS also improved flexibility over the Baseline, because the LLM has more knowledge about different domains, while SPECTER was specific for the dataset (used for training) covering the medical/biological domain; note that SPECTER-AllenAI can be considered a very strong baseline for the requested task, as it is specifically designed for scientific literature.
 
\subsection*{Dashboard for interactive exploration}

The results of the TETYS pipeline are made available through a Web application that allows users to appreciate the topics (resulting from the topic modeling sub-pipeline) and their characteristics, including their representation in time (resulting from the topic exploration sub-pipeline).

\begin{figure} [h!]
    \centering
   \includegraphics[width=\linewidth]{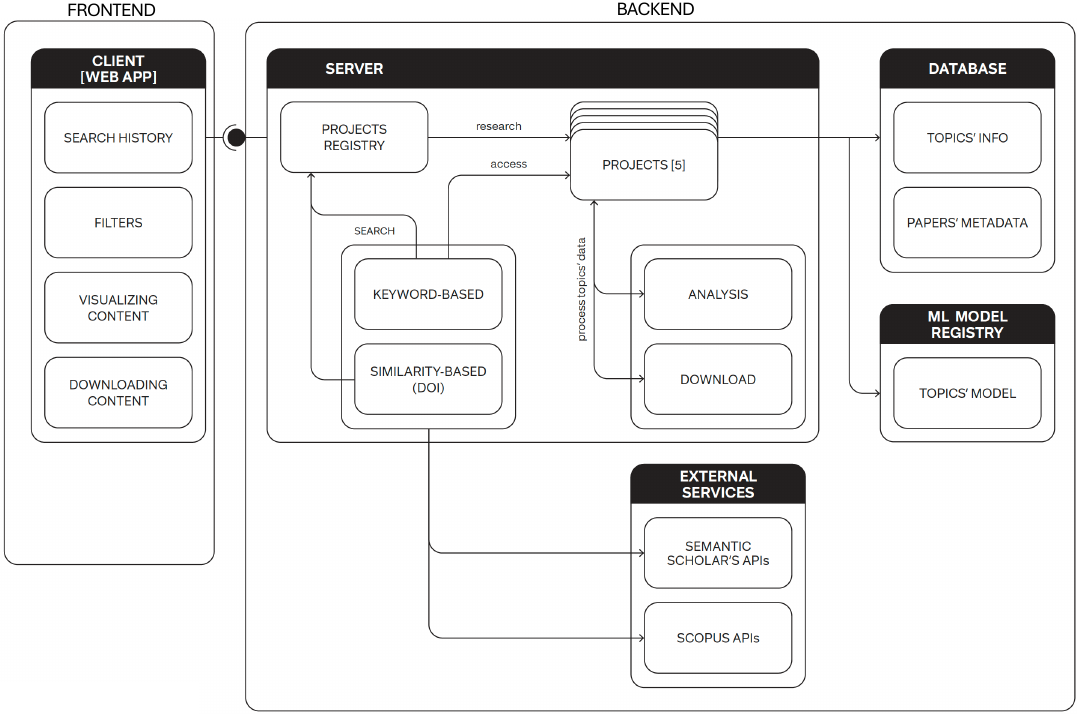}
    \caption{System architecture}
    \label{fig:informationarchitecture}
\end{figure}

Figure~\ref{fig:informationarchitecture} represents the system architecture divided into a \textit{frontend} and a \textit{backend}.
The frontend contains a Web application working as a \textbf{Client} with functionalities that allow users to select a macro-area of interest, filter the content of the topic model using keywords or a specific publications' DOI, visualize the content, and download it (through plots and tables).

The backend contains four modules. 
Data persistence is taken care of in the 
\textbf{Database} (collecting publications metadata and information describing the topics, like their trends over time, stored as time-series data) and in the 
\textbf{ML Model registry}, which stores the topics models of the project as large \textit{pickle} objects.
The database is implemented with DuckDB \cite{raasveldt2019duckdb}, an in-process analytical database, that we use to exploit the efficiency in data storage and retrieval of the Apache Parquet format \cite{apacheparquet}.
These two modules can be queried by the central \textbf{Server}, i.e., the orchestrator of TETYS: this includes a project registry along with services to perform keyword-based search and similarity-based search over the five different projects (one per macro-area), which continue to send and receive data. 
In each project, we allow analysis (i.e., statistical testing) and results download. Keyword-based search is exploited to find ranked topics that are close (i.e., relevant) to specific keywords. Similarity-based search is exploited to find ranked topics that are relevant to a specific point in the embedding space, i.e., one abstract -- identified through its DOI. These search procedures make use of \textbf{External services} such as Scopus APIs~\cite{scopus_search_api} and Semantic Scholar's APIs~\cite{semscholar}.
Note that the Model registry contains the models that, for each project, infer the most relevant topics for any query, both keywords-based and DOI-based.

\begin{figure} [h!]
    \centering
   \includegraphics[width=\linewidth]{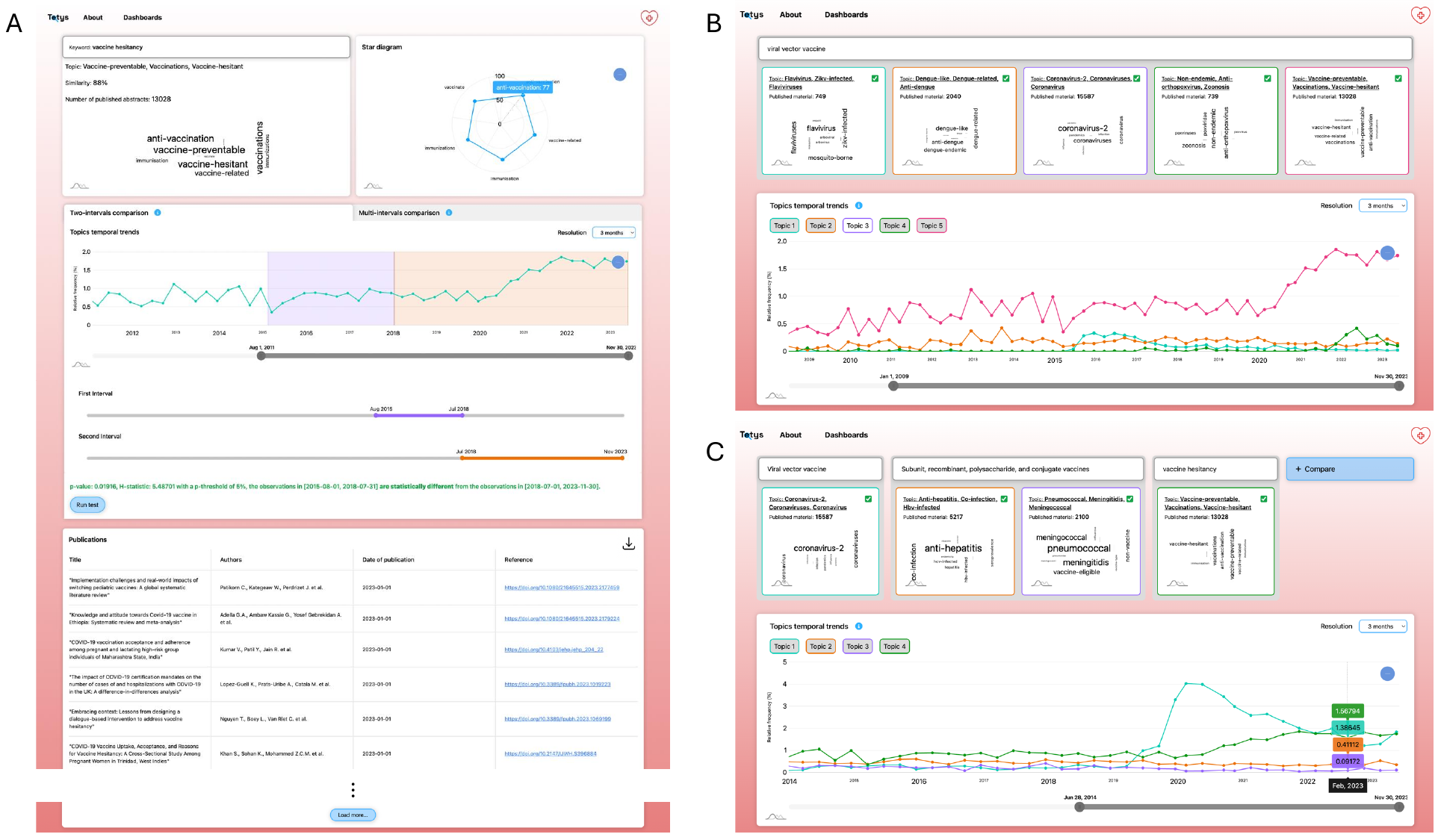}
    \caption{Pages of the TETYS Dashboard}
    \label{fig:screens}
\end{figure}

The TETYS dashboard allows us to directly inspect the results obtained by our pipeline, supporting users in the exploration of topics, which would be a tedious and time-consuming task if performed manually.
Users are asked to select one macro-area out of the five offered. For each macro-area, they can either select one of the trending topics shown in a scrollable gallery or start their search using a keyword.
These two possibilities allow them to access two possible pages:
the \textit{Single Topic} page (see Figure~\ref{fig:screens}, Panel~A) or the \textit{Topic Comparison} page (see Figure~\ref{fig:screens}, Panels~B/C).
Panel~A shows a descriptive card of the topic with its wordcloud and star diagram, a component for performing two-interval or multi-interval comparisons between user-selected time spans of the topic time series, and a downloadable list of publications that are assigned to the topic.
Panel~B shows a set of topics selected by the user from a pool of topics related to the searched keyword; topics (max. 5) can be selected also during multiple consecutive searches (as shown in Panel~C). 
Their corresponding time series are shown on the same graph, where users can (de)select tracks as needed and use a slider to focus on a time span of interest.
Different time resolutions can be set; the relative frequencies of the topics in one specific time instant can be visualized on hover.

%\subsubsection*{Use case}
%\subsubsection*{User Evaluation}

\section*{Discussion}
\label{sec:discussion}

The proposed system presents a series of innovations that include the possibility of applying the BERTopic pipeline in a customized way on big data corpora, the optimization of the hyperparameter search, and the storage of intermediate models to obviate the stochastic nature of HDBSCAN.
From the technological point of view, our system poses the basis for applying the pipeline to many diverse domains and text corpora, provided that the constraints of our setup are observed.

We observed that BERTopic models developed with LLM-based embedding models typically identified more topics than models developed with non-LLM-based embedding models. One of the likely reasons is the dimensionality of the embedding vectors, which is much larger in the case of LLM-based embedding models (4096 $>>$ 768). 
In a larger latent space, the model has a better capacity to distinguish between similar but different topics, which can be difficult for models in a small latent space. In addition to the larger dimensionality of the latent space and better semantic representation, LLM-based embedding models are, in their essence, more powerful, since they are pre-trained on much larger and extensive text data, on top of using more advanced learning techniques and fine-tuning.

A limitation in the current approach is related to the representation of topics.
Since we run topic modeling as an unsupervised task on a high-dimensional latent space, given topics may appear not to be precisely separated from a textual perspective -- as they can share terms in their representations. 
Through manual investigation, we verified that this is not due to limitations in the topics' identification process;
instead, the problem rather pertains to representation extraction.
We are confident that this issue will be solved with the application of new language models that are fine-tuned for this purpose.

Moreover, in our evaluation, we did not discuss stability and efficiency~\cite{abdelrazek2023topic} of our topic model, as they are not integral to our process.
Note that, after the first fitting of the topic model, we reuse the model and update it with new entries, during inference, without being affected by concerns of stability or efficiency.

Regarding the specific working instance exposed in the TETYS dashboard, focusing on SDGs-related literature, we believe the system can be useful to a very broad range of stakeholders, including users such as students, researchers, or professionals who are interested in deepening their knowledge on an area of research and need a fast way to grasp a general idea of the main topics and their evolution in the last twenty years.
Possibly, one such dashboard could be extended into a product useful to funding bodies, universities, or research centers.

%Implications for various stakeholders 
%(see~\cite{raman2024green}

%\subsection*{Qualitative evaluation}
%\cite{abdelrazek2023topic}

%- Guardare come discutere topic
%quality (model choice+dim.red.)
%interpretability (model choice+dim.red.)
%stability NON VALUTABILE
%diversity (model choice+dim.red.)
%efficiency (random search vs grid search)
%flexibility (hyperparam opt)

\section*{Conclusion}
\label{sec:conclusion}

The TETYS pipeline is based on BERTopic; we enhanced it by using LLMs for the embedding computation. Then, for each data corpus at hand, we can find a local maximum in the random search space of the hyperparameter configuration that regards dimensionality reduction and clustering.
This configuration is used for model registration and fitting.
Given a corpus of text documents in input, eventually, our pipeline builds a valuable trade-off between the best and ``fastest-to-find'' topic model possible. 
We measure the goodness of configurations one by one by leveraging DBCV, while we assess the overall arrangement with a thorough manual evaluation.

This arrangement is particularly fit for big data corpora; we additionally enrich the pipeline result by enabling keyword-search and dynamic topic modeling with time series exploration using configurable time-bins and relative frequencies (with ranking).
The final result exposes a rich computational model and associated metadata to the users, making topics' exploration interactive and possible on a large scale.

In this work, we demonstrated the power of the TETYS pipeline by running it on five different text document corpora generated from the Scopus database by collecting research abstracts that are pertinent to a specific set of Sustainable Development Goals, as defined by the United Nations.
This approach allowed us to automatically uncover the attitude and the topic trends found in research literature about themes of interest for the general community.

Moving beyond research-related text, we expect to re-apply the pipeline and its paradigm to other application domains that may particularly benefit from this kind of data analytics, i.e., analyzing topics' evolution of legislative text from different countries and systems.

\newpage

\begin{appendices}
\section*{Appendix}
\input{parameters_table}    
\end{appendices}

\newpage

\section*{Declarations}

\bmhead{List of abbreviations}
\hfill \break
\noindent
DBCV: Density-Based Clustering Validation\\
DOI: Digital Object Identifier\\
HDBSCAN: Hierarchical Density-Based Spatial Clustering of Applications with Noise\\
LLM: Large Language Model\\
MMR: Maximal Marginal Relevance\\
MTEB: Massive Text Embedding Benchmark\\
SBERT: Sentence Bidirectional Encoder Representations from Transformers\\
SDG: Sustainable Development Goal\\
SPECTER: Scientific Paper Embeddings using Citation-informed TransformERs\\
TETYS: Topics Evolution That You See\\
UMAP: Uniform Manifold Approximation and Projection

\bmhead{Ethics approval and consent to participate}
Not applicable.

\bmhead{Consent for publication}
Not applicable.

\bmhead{Availability of data and materials}
The pipeline is packaged as a Python module and organized as a series of scripts. The code is available on the GitHub repository \url{https://github.com/FrInve/TETYS}, under the BSD-3-Clause license; it contains requirements to reproduce the conda environment and code/data of intermediate steps.\\
The web application is freely available in the same GitHub repository.%at \url{https://geco.deib.polimi.it/tetys} and the RESTful APIs are available at \url{https://geco.deib.polimi.it/tetys_api/}.
%The software is completely Open Source, in line with the principles of the NGI Search Initiative. 

\bmhead{Competing interests}
The authors declare that there are no conflicts of interest.

\bmhead{Funding}
\hfill \break
\noindent
\begin{minipage}
{0.78\textwidth}
This research is supported by the TEThYS project, a beneficiary of the NGI Search 2nd Open Call (Number of the Sub-grant Agreement SEARCH OC2\_18), funded by the European Union. The NGI Search project has received funding from the European Union’s Horizon Europe research and innovation programme under the grant agreement 
101069364 and it is framed under the Next Generation Internet Initiative
\end{minipage}
\hfill%
\begin{minipage}
{0.20\textwidth}% adapt widths of minipages to your needs
\includegraphics[width=0.8\linewidth]{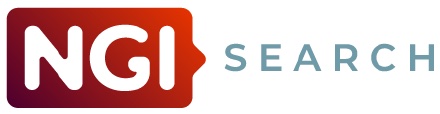}
\includegraphics[width=0.8\linewidth]{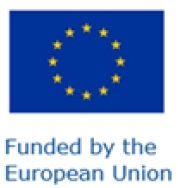}
\end{minipage}%

\vspace{3mm}
\bmhead{Authors' contributions}
F.I. and A.B. jointly conceptualized the work and its methodology and revised/edited the manuscript;
F.I. designed and implemented the data pipeline, the backend, and the frontend, while supervising the work of F.C., J.J., and A.S.;
F.C. contributed to the User interface and UX design;
J.J. contributed to the datasets' preparation and the topic modeling/exploration pipeline and ran the inference evaluation;
A.S. contributed to the frontend of the dashboard;
A.B. supervised the team, acquired funding, coordinated the project, and wrote the original draft.

\bmhead{Acknowledgements}
The authors are grateful to Stefano Ceri, Lorenzo Mari, and Mark J. Carman for the useful advice during the course of the project.

%\bmhead{Authors' information}

\newpage

\bibliography{sn-bibliography}% common bib file

\end{document}

%% file: parameters_table.tex
\begin{table}[h!]
\centering
\tiny
\begin{tabular}{ccclc}
\textbf{Macro area} & \textbf{Config.} & \textbf{Pipeline step} & \textbf{Parameter} & \textbf{Value} \\ \midrule
\multirow{10}{*}{M1} & \multirow{5}{*}{\makecell{Baseline\\301 topics}} & \multirow{3}{*}{UMAP} & n\_neighbors & 20 \\ 
%\cline{4-5}
                                  &&& n\_components & 5 \\ %\cline{4-5}
                                  &&& min\_dist & 0.0 \\ %\cline{4-5}
                                   \cline{3-5}
&&\multirow{2}{*}{HDBSCAN} & min\_cluster\_size & 25 \\ 
%\cline{4-5}
                                     &&& min\_samples & 100 \\ %\cline{4-5}%\cline{4-5}
                                      \cline{2-5}

&\multirow{5}{*}{\makecell{TETYS\\550 topics}} & \multirow{3}{*}{UMAP} & n\_neighbors & 20 \\ 
%\cline{4-5}
                                  &&& n\_components & 5 \\ %\cline{4-5}
                                  &&& min\_dist & 0.0 \\ %\cline{4-5}
                                   \cline{3-5}
&&\multirow{2}{*}{HDBSCAN} & min\_cluster\_size & 25 \\ 
%\cline{4-5}
                                     &&& min\_samples & 75 \\ %\cline{4-5}
                                     %\cline{4-5}
                                     \cline{1-5}

\multirow{10}{*}{M2} &\multirow{5}{*}{\makecell{Baseline\\424 topics}} & \multirow{3}{*}{UMAP} & n\_neighbors & 50 \\ %\cline{4-5}
                                  &&& n\_components & 10 \\ %\cline{4-5}
                                  &&& min\_dist & 0.0 \\ %\cline{4-5}
                                   \cline{3-5}
&&\multirow{2}{*}{HDBSCAN} & min\_cluster\_size & 25 \\ %\cline{4-5}
                                     &&& min\_samples & 50 \\ %\cline{4-5}
                                     %\cline{4-5}
                                     \cline{2-5}

&\multirow{5}{*}{\makecell{TETYS\\856 topics}} & \multirow{3}{*}{UMAP} & n\_neighbors & 20 \\ %\cline{4-5}
                                  &&& n\_components & 10 \\ %\cline{4-5}
                                  &&& min\_dist & 0.0 \\ %\cline{4-5}
                                   \cline{3-5}
&&\multirow{2}{*}{HDBSCAN} & min\_cluster\_size & 25 \\ %\cline{4-5}
                                     &&& min\_samples & 75 \\ %\cline{4-5}
                                      %\cline{4-5}
                                     \cline{1-5}

\multirow{10}{*}{M3} &\multirow{5}{*}{\makecell{Baseline\\98 topics}} & \multirow{3}{*}{UMAP} & n\_neighbors & 20 \\ %\cline{4-5}
                                  &&& n\_components & 10 \\ %\cline{4-5}
                                  &&& min\_dist & 0.0 \\ %\cline{4-5}
                                   \cline{3-5}
&&\multirow{2}{*}{HDBSCAN} & min\_cluster\_size & 25 \\ %\cline{4-5}
                                     &&& min\_samples & 50 \\ %\cline{4-5}
                                      %\cline{4-5}
                                     \cline{2-5}

&\multirow{5}{*}{\makecell{TETYS\\181 topics}} & \multirow{3}{*}{UMAP} & n\_neighbors & 100 \\ %\cline{4-5}
                                  &&& n\_components & 10 \\ %\cline{4-5}
                                  &&& min\_dist & 0.0 \\ %\cline{4-5}
                                 \cline{3-5}
&&\multirow{2}{*}{HDBSCAN} & min\_cluster\_size & 25 \\ %\cline{4-5}
                                     &&& min\_samples & 10 \\ %\cline{4-5}
                                     %\cline{4-5}
                                    \cline{1-5}

\multirow{10}{*}{M4} &\multirow{5}{*}{\makecell{Baseline\\42 topics}} & \multirow{3}{*}{UMAP} & n\_neighbors & 50 \\ %\cline{4-5}
                                  &&& n\_components & 28 \\ %\cline{4-5}
                                  &&& min\_dist & 0.0 \\ %\cline{4-5}
                                   \cline{3-5}
&&\multirow{2}{*}{HDBSCAN} & min\_cluster\_size & 25 \\ %\cline{4-5}
                                     &&& min\_samples & 50 \\ %\cline{4-5}
                                     %\cline{4-5}
                                    \cline{2-5}

&\multirow{5}{*}{\makecell{TETYS\\136 topics}} & \multirow{3}{*}{UMAP} & n\_neighbors & 50 \\ %\cline{4-5}
                                  &&& n\_components & 28 \\ %\cline{4-5}
                                  &&& min\_dist & 0.0 \\ %\cline{4-5}
                                   \cline{3-5}
&&\multirow{2}{*}{HDBSCAN} & min\_cluster\_size & 25 \\ %\cline{4-5}
                                     &&& min\_samples & 10 \\ %\cline{4-5}
                                      %\cline{4-5}
                                     \cline{1-5}

\multirow{10}{*}{M5} &\multirow{5}{*}{\makecell{Baseline\\291 topics}} & \multirow{3}{*}{UMAP} & n\_neighbors & 5 \\ %\cline{4-5}
                                  &&& n\_components & 5 \\ %\cline{4-5}
                                  &&& min\_dist & 0.0 \\ %\cline{4-5}
                                  \cline{3-5}
&&\multirow{2}{*}{HDBSCAN} & min\_cluster\_size & 25 \\ %\cline{4-5}
                                     &&& min\_samples & 10 \\ %\cline{4-5}
                                      %\cline{4-5}
                                     \cline{2-5}

&\multirow{5}{*}{\makecell{TETYS\\167 topics}} & \multirow{3}{*}{UMAP} & n\_neighbors & 100 \\ %\cline{4-5}
                                  &&& n\_components & 35 \\ %\cline{4-5}
                                  &&& min\_dist & 0.0 \\ %\cline{4-5}
                                   \cline{3-5}
&&\multirow{2}{*}{HDBSCAN} & min\_cluster\_size & 25 \\ %\cline{4-5}
                                     &&& min\_samples & 15 \\ %\cline{4-5}
                                     %\cline{4-5}
                                      \cline{1-5}
\end{tabular}
\caption{Hyperparameter values for macro areas M1 to M5, in both configurations Baseline (non-LLM with grid search) and TETYS (LLM with random search), considering the UMAP and HDBSCAN steps.
From both steps, we omit \texttt{metric = `euclidean'} as it is always the same value for both UMAP and HDBSCAN, as well as \texttt{cluster\_selection\_method = `eom'} as it is always the same value for HDBSCAN.}
\label{table:parameters}
\end{table}